\documentclass[letterpaper, 10 pt, conference]{ieeeconf}
\IEEEoverridecommandlockouts
\usepackage[space, compress, sort]{cite}
\usepackage{amsmath}
\usepackage{amsfonts}

\usepackage{amsthm, amssymb}
\usepackage[ruled,vlined]{algorithm2e}
\usepackage{mathtools}
\usepackage{tabularx}
\usepackage{graphicx}
\usepackage{enumerate}
\usepackage{float}
\usepackage{url}
\usepackage{verbatim}
\usepackage{booktabs} 
\usepackage{multirow}
\usepackage[linkcolor=black,citecolor=black,urlcolor=black,colorlinks=true]{hyperref}
\usepackage{bm}
\usepackage{subfigure}
\usepackage{algpseudocode}

\newtheorem{definition}{Definition}

\newcommand{\delete}[1]{{\bgroup\markoverwith{\textcolor{red}{\rule[0.5ex]{2pt}{0.4pt}}}\ULon{#1}}}

\newcommand{\tabincell}[2]{\begin{tabular}{@{}#1@{}}#2\end{tabular}}

\title{\LARGE \bf Safe Interval Motion Planning for Quadrotors in Dynamic Environments}
\author{Songhao Huang$^{*}$, Yuwei Wu$^{*}$, Yuezhan Tao, Vijay Kumar
\thanks{$^{*}$Equal contribution. The authors are with the GRASP Laboratory, University of Pennsylvania, Philadelphia, PA, 19104 USA {\tt\small\{songhaoh, yuweiwu, yztao, kumar\}@seas.upenn.edu}.}%
\thanks{This research was sponsored by TILOS under NSF grants CCR-2112665.}
}
\begin{document}
\maketitle

\begin{abstract}
Trajectory generation in dynamic environments presents a significant challenge for quadrotors, particularly due to the non-convexity in the spatial-temporal domain.
Many existing methods either assume simplified static environments or struggle to produce optimal solutions in real-time.
In this work, we propose an efficient safe interval motion planning framework for navigation in dynamic environments.
A safe interval refers to a time window during which a specific configuration is safe.
Our approach addresses trajectory generation through a two-stage process: a front-end graph search step followed by a back-end gradient-based optimization.
We ensure completeness and optimality by constructing a dynamic connected visibility graph and incorporating low-order dynamic bounds within safe intervals and temporal corridors.  
To avoid local minima, we propose a Uniform Temporal Visibility Deformation (UTVD) for the complete evaluation of spatial-temporal topological equivalence.
We represent trajectories with B-Spline curves and apply gradient-based optimization to navigate around static and moving obstacles within spatial-temporal corridors.
Through simulation and real-world experiments, we show that our method can achieve a success rate of over 95\% in environments with different density levels, exceeding the performance of other approaches, 
demonstrating its potential for practical deployment in highly dynamic environments.

\end{abstract}

\IEEEpeerreviewmaketitle
\vspace{-0.1cm}
\section{Introduction}

%%%%%%%%%%part one: general problem for obstacle avoidance

Trajectory generation for autonomous navigation in static environments has been widely applied in various fields, including forestry, industry, and agriculture~\cite{https://doi.org/10.1002/rob.20403, 10.1007/978-3-030-33950-0_17, 9720974}.
However, the assumption of a static or nearly static environment may not always hold, especially in urban low-altitude scenarios. 
Previous works adhered to the static assumption and triggered replanning whenever the map was updated~\cite{1013481, KOENIG200493}.
Despite increasing the frequency of an online replanning framework, finding a feasible trajectory remains challenging in such dynamic environments.

%%%%%%%%%%%part two: narrow the gap
A straightforward method for extending motion planning algorithms from static to dynamic environments involves introducing an additional time 
dimension~\cite{doi:10.1177/0278364915614386, 5152860, topo1, 1642056}.
However, this significantly expands the state space, leading to redundant map traversal and making online solutions challenging in the presence of moving obstacles.
Traditional approaches mainly focus on reducing the search space by finding safe intervals in planning and searching paths as an initial guess for trajectory generation, aiming to find solutions within limited time budgets~\cite{SIPP1, anytimeSIPP, lazysipp, kinoSIPP, 8740885}.
% However, no complete and principal way ensures the dynamic feasibility of high-order systems while remaining efficient for onboard planning in dynamic environments.
However, there is no comprehensive and fundamental method that ensures the dynamic feasibility of high-order systems while remaining efficient for onboard planning in dynamic environments.

\begin{figure}[!t]
      \vspace{0.1cm}
      \centering
      \includegraphics[width=0.95\columnwidth]{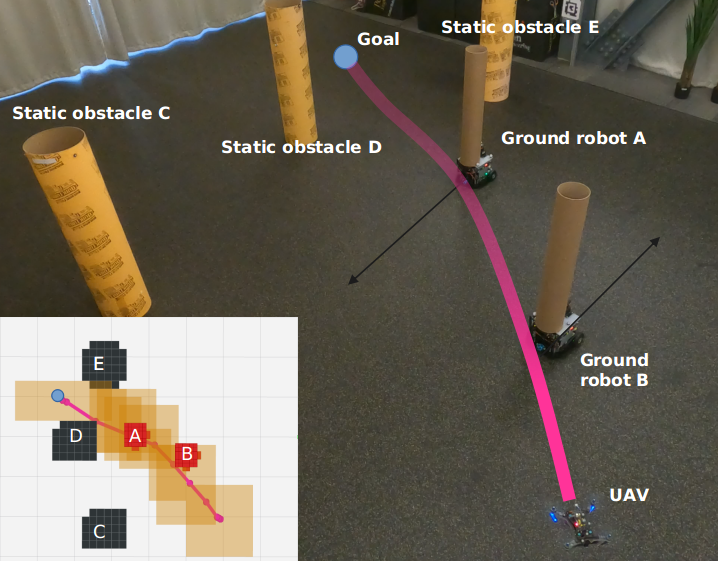}
      \vspace{-0.25cm}
      \caption{A representative experiment of dynamic obstacle avoidance. The quadrotor plans a trajectory (purple) and navigates between static obstacles (cylinders) and moving obstacles (ground robots).}
      \label{fig:fig_env_show}
      \vspace{-0.8cm}
\end{figure}

To address these issues, we propose a complete dynamic connected visibility graph construction method and evaluate path equivalence using the concept of \textit{Uniform Temporal Visibility Deformation} to get multiple topological distinct initial paths.
To incorporate robot dynamics into the graph construction process, we employ the double integrator model for velocity profile to approximate minimum travel durations. 
This ensures the lower-order dynamic feasibility of safe intervals and temporal corridors while providing sufficient flexibility for higher-order adjustments during the back-end optimization.
To ensure smoothness, B-spline trajectories are then optimized within the spatial-temporal corridors generated from initial paths.
The trajectory with the minimum control effort is selected.
We analyze that our method is probabilistic complete and optimal given dynamic bounds.
Our contributions can be summarized as follows, 
\vspace{-0.1cm}
\begin{itemize}
    \item We propose a safe interval motion planning framework for dynamic environments consisting of front-end topological path searching and back-end optimization using spatial-temporal corridors.
    \item We introduce a dynamic connected visibility graph construction method with guarantees of second-order dynamic feasibility. We define \textit{Uniform Temporal Visibility Deformation} (UTVD) to evaluate the spatial-temporal topological equivalence.
    % \item We demonstrate in both simulation benchmark comparison and real-world deployment on our customized hardware platforms.
    \item We conduct extensive simulation comparisons and hardware experiments to validate the effectiveness of the proposed framework.
    
\end{itemize}

\section{Related Works}

\subsection{Moving Obstacle Avoidance}

Motion planning in the presence of moving obstacles introduces challenges for finding complete and efficient strategies to resolve conflicts.
A two-stage planning framework, studied in \cite{10610207}, efficiently generates high-quality trajectories by using a front-end planner to identify collision-free paths and then optimize trajectories for smoothness, feasibility, and safety.
% To generate high-quality trajectories with limited computation times, a two-stage planning framework, extensively studied and benchmarked in~\cite{10610207}, offers an efficient pipeline to address computation constraints and improve the feasibility of planning algorithms. 
% It employs a front-end planner to identify one or multiple collision-free low-order paths, which are further optimized within the same homotopic trajectory class to ensure smoothness, feasibility, and safety.
Graph-based front-end methods like Probabilistic Roadmaps (PRM)~\cite{508439} and its variations~\cite{1041613, jaillet2008path, fasttopo} generate multiple distinct initial paths to avoid potential infeasible local minima and improve overall performance. 
With feasible initial paths, optimization-based approaches such as Model Predictive Control~\cite{MPC} or minimum control trajectory generation based on flatness properties~\cite{5980409} can be effectively employed.
In~\cite{fasttopo}, a relaxed formulation for homotopy equivalence was proposed as Uniform Visibility Deformation (UVD). 
% and further applied in~\cite{topo1} for practical evaluation for moving obstacle avoidance. 
This approach was subsequently applied in~\cite{topo1} for practical evaluation in the context of moving obstacle avoidance.
This work was further improved in~\cite{topo2} by accommodating multiple goals in Visibility-PRM and incorporating homotopy constraints into the optimization process.
% However, the equivalence evaluation remains incomplete in the temporal domain, and the feasibility of the path is not fully guaranteed.
However, the equivalence evaluation becomes incomplete when considering paths in different temporal domains, i.e., the start and end times of two paths are not the same.
To address this, we introduce a complete criterion for spatial-temporal topological evaluation to generate initial paths for further trajectory optimizations. 

\subsection{Planning with Safe Intervals}
To efficiently generate a path or trajectory in non-convex static environments,~\cite{Deits14computinglarge} proposed the use of convex decomposition for space reduction and approximation, which was further extended in~\cite{7839930} to create safe flight corridors.
To address the challenge of moving obstacle avoidance, spatial-temporal corridors~\cite{ding2019safe} have been directly generated on semantic maps with temporal information for 2-D scenarios of autonomous driving. 
Refining existing 3-D static corridors in global maps to accommodate moving objects has been explored in~\cite{9981447}, enabling the online update and reconstruction of these corridors.
However, these methods rely on the availability of feasible initial paths within the same fixed temporal corridors, lacking the flexibility to adapt to different temporal interval combinations.

The Safe Interval Path Planning (SIPP) algorithm~\cite{SIPP1} proposed a complete approach that decomposes the temporal domain into intervals on grid maps to reduce the size of the search space. 
However, evaluating time intervals for all grids is still computationally expensive and the robot dynamics are not considered in this phase. 
Various SIPP-based algorithms are subsequently introduced to solve these issues~\cite{anytimeSIPP, lazysipp, kinoSIPP}. However, the computation cost for setting up time intervals in 3D environments remains high.
To reduce the space complexity, graph structures like Probabilistic Roadmaps (PRMs) are used to represent the environment.
% Considering sampling the time in Traditional PRMs while ensuring continuity with the temporal domain
% leads to a directed graph structure since time is monotonically increasing, fundamentally changing the nature of PRMs.
Temporal PRM \cite{TPRM} applied safe intervals to PRM, reducing the complexity by only sampling vertices in static environments, and enabling multiple queries on the roadmap. 
However, this approach relies on simplistic motion assumptions.
It only evaluates the collision at vertices for efficiency by limiting the maximum length of edges.
This requires a dense graph, and the quality of paths largely depends on the maximum edge length. 
Our method addresses these challenges by incorporating safe intervals of edges and low-order dynamics in the dynamic connected visibility graph, extending distinct topological paths into spatial-temporal corridors, and further optimizing the trajectories based on flatness-based dynamics to ensure feasibility and smoothness.

\section{Prerequisites}
\subsection{Problem Formulation}
We consider a quadrotor navigating a 3D environment containing static and moving obstacles.
The environment is fully observed within a finite time range $T= [t_s, t_e]$.
It can be expressed as $\mathbb{R}^3 = \mathcal{X}_{free}^t  \cup \mathcal{X}_{obs}^t, \forall t \in T $, where $\mathcal{X}_{free}^t$ is the free space and $\mathcal{X}_{obs}^t$ is the obstacle space at any time in $T$.
The moving obstacles' trajectories are known in $T$ and have bounded dynamics.
% We consider the dynamics of the robot and obstacles as
% \begin{align}
% \left\{\begin{matrix*}[l]
% &\dot{x}(t) = f(x(t), u(t)) \\ 
% &\dot{x}_{o_i}(t) = f_{o_i}(x_{o_i}(t), u_{o_i}(t)) \\
% & \forall o_i \in \mathcal{X}_{obs}^T,   \ \ \forall t\in [t_s, t_e] \\ 
% \end{matrix*}\right.          
% \end{align}
% where $\{x(t), u(t)\}$ represents the state and input of the robot, $\{x_{o_i}(t), u_{o_i}(t)\}$ represents the state and input for obstacle $o_i$.
% The moving obstacles have bounded dynamics.
Our objective is to generate a trajectory that has minimum control cost while ensuring safety, smoothness, and dynamical feasibility from the start location to the goal region within $T$. 

% \begin{figure}[!ht]
%     \centering
%     \includegraphics[width=1.0\columnwidth]{fig/UTVD.png}

%     \vspace{-0.5cm}
%     \caption{Example of evaluation of temporal Topology Equivalence in a 2-D scene (right). The corresponding spatial-temporal topological equivalent relations for the same geometric path at x=[5, 6] (left). Red grids mean they are occupied by moving obstacles at specific time durations, colored solid line segments represent multiple paths from start to end vertex, while red dash lines represent collisions detected, otherwise they are black. According to the definition of UTVD, paths in red and blue belong to the same UTVD class. Path in green is in a different UTVD class}
%     \vspace{-0.5cm}
%     \label{fig: fig_UTVD_show}
% \end{figure}

\vspace{-0.5cm}
\begin{figure}[!ht]
    \centering
    \subfigure[Example of evaluation of spatial-temporal topological equivalence in a 2-D scene.]{
\includegraphics[width=0.46\columnwidth]{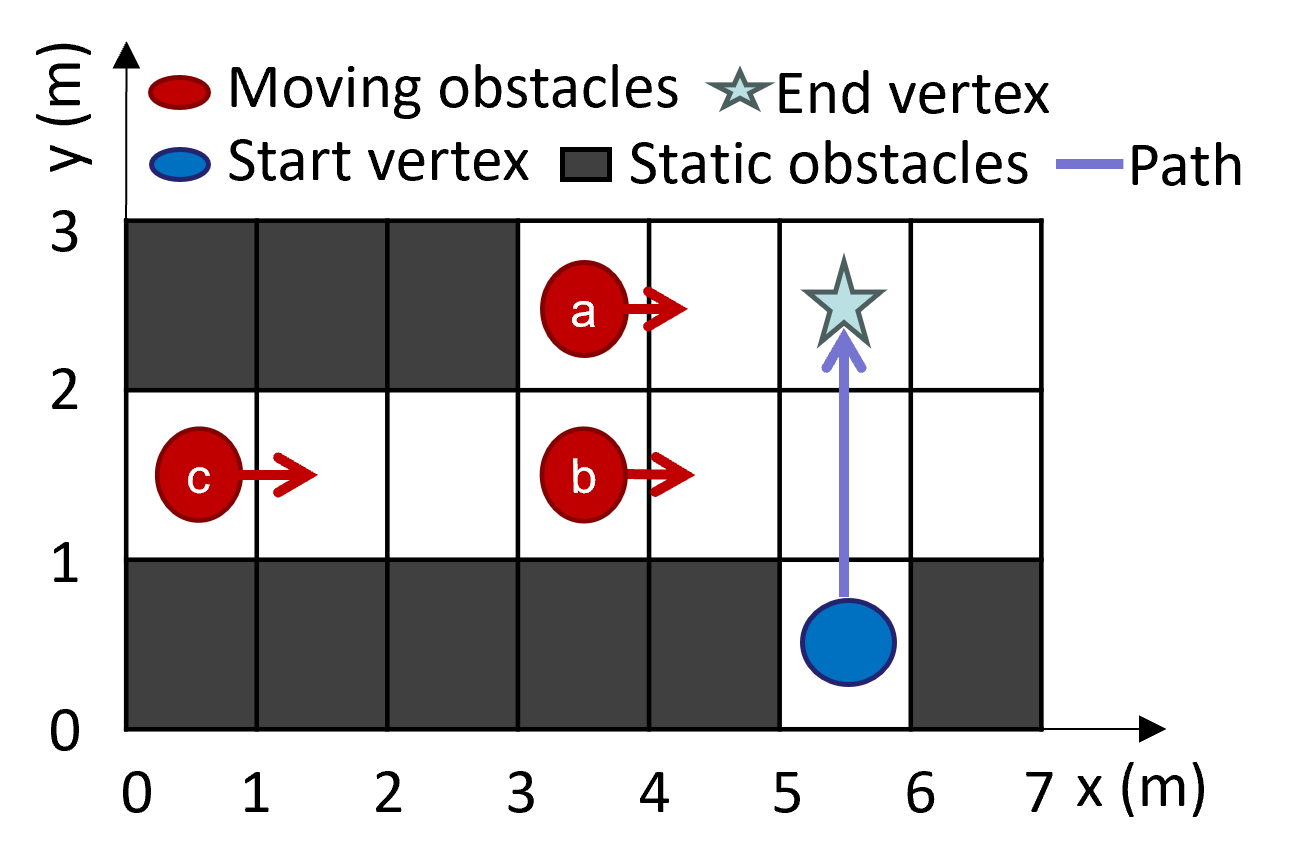}
      \label{fig:UTVD2}
    }
    \subfigure[Spatial-temporal topological equivalence for the same geometric paths at x = \( \text{[5, 6]}\).]{
\includegraphics[width=0.46\columnwidth]{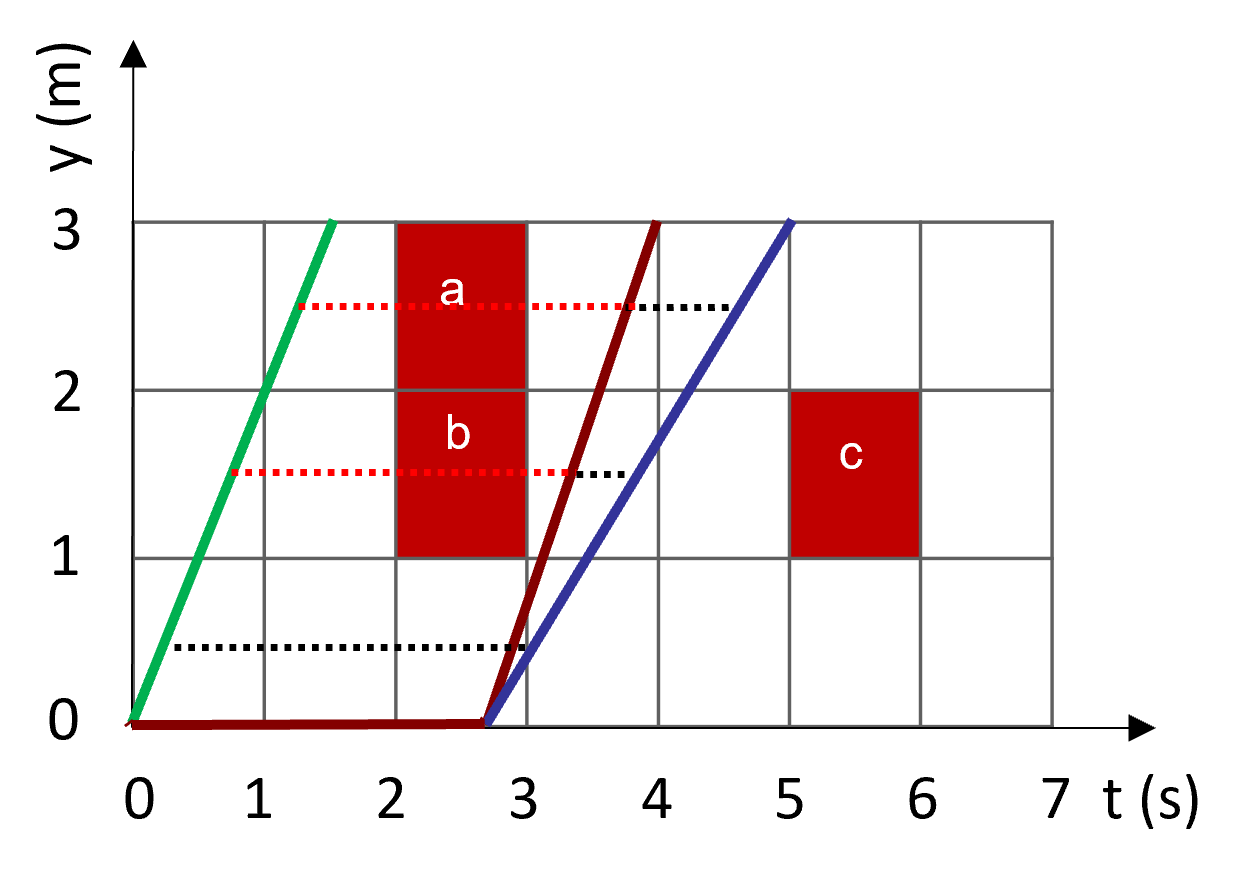}
        \label{fig:UTVD1}
    }
    \vspace{-0.5cm}
    \caption{In this example, red grids are occupied by moving obstacles at specific time durations, colored solid line segments represent multiple paths from start to end vertex, while red dash lines represent collisions detected, otherwise, they are black. According to the definition of UTVD, paths in red and blue belong to the same UTVD class. The path in green is in a different UTVD class.}
    \vspace{-0.5cm}
 \label{fig: fig_UTVD_show}
\end{figure}

\subsection{Temporal Topology Equivalence}

The geometrically topological complexity of the environment can be captured by the number of homotopy or homology classes of trajectories \cite{Munkres1974TopologyAF, bhattacharya2010search}.
One relation is to check the visibility deformation (VD), proposed in \cite{jaillet2008path}.
The Uniform Visibility Deformation (UVD)~\cite{fasttopo} employs VD's subset to evaluate the equivalence in the static environment efficiently.
% \begin{definition} (Uniform Visibility Deformation)  \cite{fasttopo}  Two trajectories $\sigma_1(s), \sigma_2(s)$, parameterized by $s \in [0, 1] $, and satisfying  $\sigma_1(0)= \sigma_2(0)$, $\sigma_1(1)= \sigma_2(1)$, belong to the \underline{same uniform} visibility deformation class, if for all s, line $\sigma_1(s)\sigma_2(s)$ is collision-free.
% \end{definition}
In dynamic environments, using UVD in time-extended state space\cite{topo1} cannot be directly applied since trajectories (paths) are within different temporal domains, and it also may result in time resolution incompleteness.
Therefore, we introduce an enhanced version of UVD, namely Uniform Temporal Visibility Deformation (UTVD), to capture trajectories in different spatial-temporal Topological classes.
\begin{definition} (Uniform Temporal Visibility Deformation (UTVD)) Two trajectories $\sigma_1(s), \sigma_2(s')$, parameterized by $s \in [0, 1] $, $s' = \alpha s + \theta, \alpha \in \mathbb{R}^+, \theta \in \mathbb{R}$, and satisfying  $\sigma_1(0)= \sigma_2(\theta)$, $\sigma_1(1)= \sigma_2(\alpha + \theta)$, belong to the \underline{same uniform temporal visibility deformation} (UTVD) class if for all s, line segment $\sigma_1(s)\sigma_2(\alpha s + \theta)$ is collision-free during $[s, \alpha s + \theta]$.
\end{definition}

We propose an exact approach to perform collision checking for lines within specific temporal Intervals (see section IV.A).  
Fig. \ref{fig: fig_UTVD_show} showcases a simple example of UTVD evaluation. 
There are three moving obstacles with speeds of $1m/s$ in the environment. 
A robot can travel along the same geometrical path at x=[5,6] safely with three paths in different temporal domains, the red and green paths belong to the different UTVD classes because collisions with obstacle a and obstacle b are detected in the grids (x=[5,6], y=[2,3]) and (x=[5,6], y=[1,2]) within temporal interval t=[2,3], respectively.
UTVD can help us evaluate the paths within different safe intervals.

\subsection{Safe Intervals and Temporal Corridors}
Two vertices can be connected by an edge in a graph.
The quadrotor can travel along the edge, where the minimum travel duration $t_{min}$ can be practically computed using a trapezoidal velocity profile.
Here we assume the quadrotor travels on edges with rest start and end states and approximate the duration accordingly to provide a lower bound of traveling time, which is practical to provide a feasible heuristic for higher-order systems~\cite{7839930}.
We now introduce the definition of safe interval for edge as

\begin{definition} (Safe Intervals for Edge) 
For an edge $e$ in the time range $T$, The Collision intervals ${\rm CI(e)}$ is defined as a series of time intervals $\bigcup_i (t_i, t_{i+1})$, where $e \cap \mathcal{X}_{obs}^{t} \neq \emptyset$ for $ \forall t \in {\rm CI}$.
Subsequently, the Safe intervals ${\rm SI(e)}$ are a series of time intervals $\bigcup_j (t_j, t_{j+1})$ that satisfy $(t_j, t_{j+1}) \in T \backslash {\rm CI}(e) \cap (t_{j+1} - t_{j} > t_{min})$.
\label{def: safeinterval}
\end{definition}

Note that safe intervals for vertices can be defined similarly. 
In our method, we focus on the safe intervals of edges, as the safe intervals of edges are subsets of the safe intervals of vertices, thus providing safer solutions if the lengths of the edges in the graph are larger than the dimensions of the obstacles.
Given a path consists of consecutive edges with safe intervals, its temporal corridor can be defined as
\begin{definition} (Temporal Corridor for Path) 
A temporal corridor for a path is a series of time intervals where the safe intervals of any two consecutive edges in the path overlap.
\label{def: tempcorr}
\end{definition}
By Definition~\ref{def: tempcorr}, temporal corridors are directed. 
In this work, we consider both directions of temporal corridors for a complete evaluation (see Algo. \ref{alg: graph}) for spatial-temporal topological equivalence.

\section{Spatial-Temporal Topological Path Planning}

\subsection{Graph Construction with Safe Intervals}

\subsubsection{Safe Intervals Generation for Edges}
We represent the static environment with a 3D occupancy map and dynamic obstacles as bounded ellipsoids with trajectories parameterized by polynomials. 
For each edge in the graph, a cuboid covering it is generated and inflated by a margin equal to the Minkowski sum of the quadrotor and moving obstacles.
Collision time stamps are determined by solving the points of intersection between the hyperplanes of the cuboid and moving obstacles' trajectories.
Finally Definition \ref{def: safeinterval} is applied for calculating the safe intervals.

% \begin{algorithm}
%     \caption{findSafeIntervals}
%     \label{alg: SI}
%     KwIn{point set $p$ , time range $T= [t_s, t_e]$, obstacles $\mathcal{X}_{obs}^T =  O_{sta} \cup O_{dyn}$, robot radius $r$}
%     \If{$p \in O_{sta}$ }
%     {
%         SI $\leftarrow \varnothing $ \\
%         \Return SI
%     }
%     SI $\leftarrow \{t_s, t_e\}$, CI $\leftarrow \varnothing $ \\
%     \For{$o_i \in O_{dyn}$}
%     {    
%         // get moving obstacle's trajectory $\xi_i(t)$ , size $l_i$ \\ 
%         $\xi_i(t), l_i \leftarrow $ \textbf{getObsInfo} ($o_i$)  \\
%         // solve intersection time stamps \\
%         $ \tau_c$ = \textbf{solve($l_i , r, \xi_i(t), p$)} \\
%         // update collision intervals  \\
%         CI $\leftarrow$ \textbf{update($\tau_c$)} 
%     }
%     // complement the collision intervals \\
%     SI $\leftarrow$ \textbf{buildSafeInterval}(CI)\\
%     \Return SI
% \end{algorithm}
% \begin{equation}
% \label{eq:inter-agent-checking}
% \sigma(\tau-\tau_p)  \cap \bigcap_{u=1}^{U} [\xi_o(\tau - \tau_o) \oplus \xi_{e}(\Phi^{u}(\tau - \tau_o))]  = \emptyset \\
% \end{equation}

\subsubsection{Dynamic Connected Visibility Graph Construction}
\begin{figure}[!ht]
    \centering
    \subfigure[Dynamic connected graph]{
      \includegraphics[width=0.46\columnwidth]{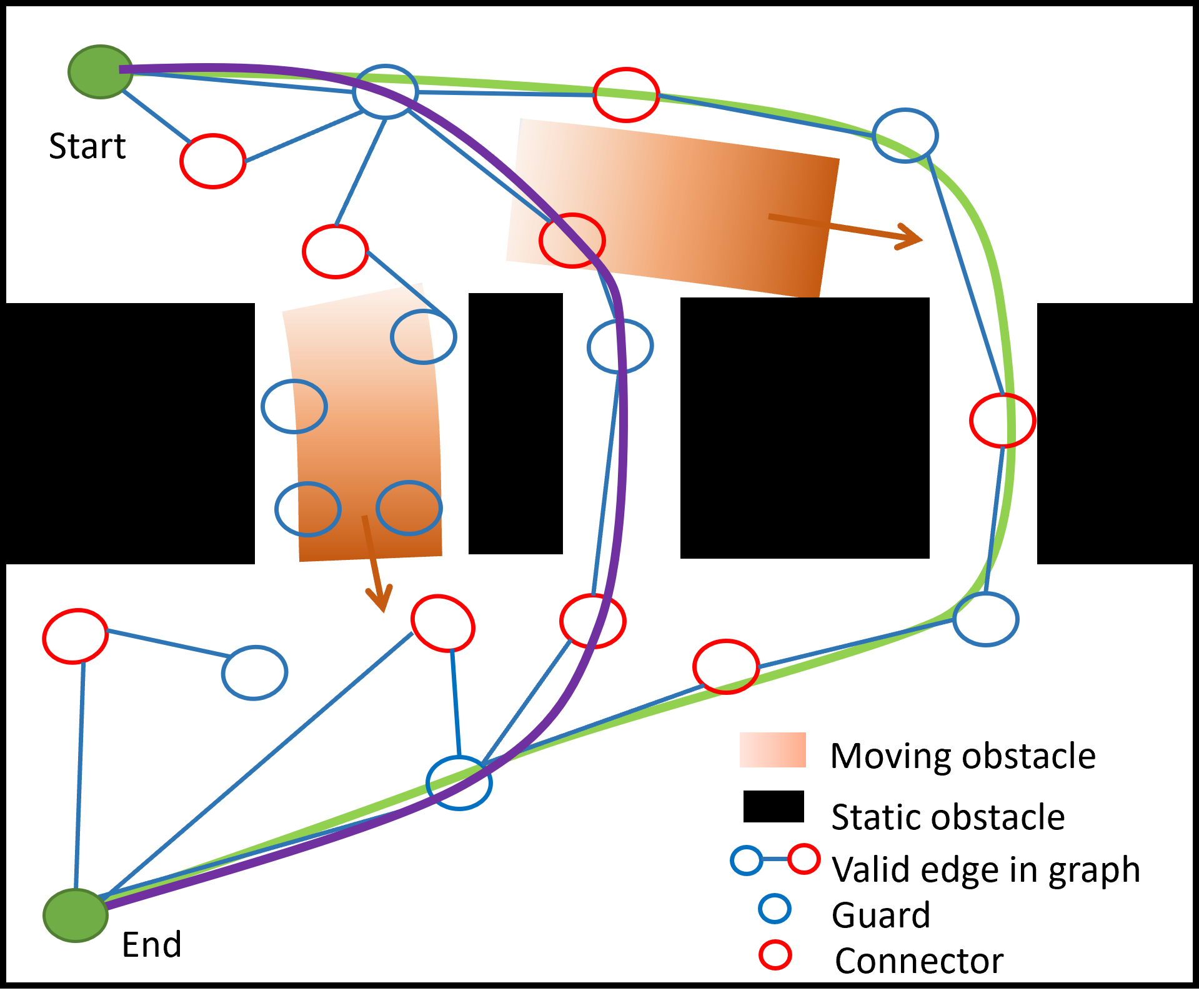}
      \label{fig:valid_graph}
    }
    \subfigure[Simulation]{
      \includegraphics[width=0.46\columnwidth]{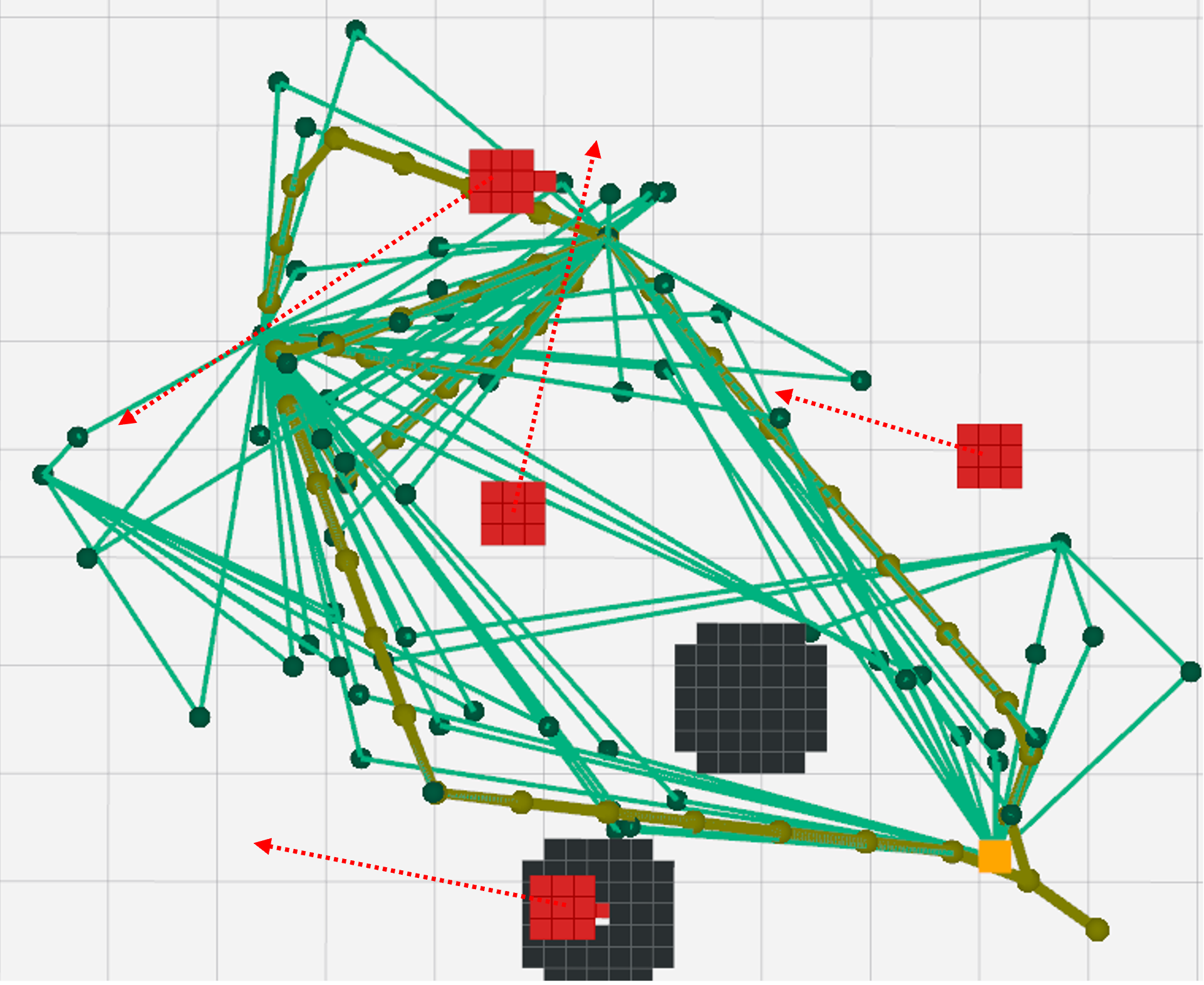}
      \label{fig:valid_graph2}
    }
    \caption{(a) Illustration of dynamically connected visibility graphs in dynamic environments. A dynamic connected graph has edges that are valid in their safe intervals. Valid paths in distinct UTVD classes are shown in purple and green. (b) Simulation result of the front-end graph search, the red objects are moving obstacles with trajectories in dash lines, and the black ones are static obstacles. The graph is shown in green, and multiple distinct paths are represented in yellow. }
    \vspace{-0.3cm}
 \label{fig:approximation}
\end{figure}

To reduce space complexity, ensure the safety on edges, and preserve the multi-query property within a given time horizon, we generate a dynamic connected visibility graph as shown in Fig. \ref{fig:approximation}.

% Traditional Probabilistic Roadmaps (PRMs) operate under the assumption that the environment remains static, making it challenging to adapt them directly for dynamic environments.
% Considering time domain into the PRMs while ensuring continuity with the temporal domain %(satisfying $\frac{\partial s}{\partial t} > 0$) %
% leads to a directed graph structure since time is monotonically increasing, fundamentally changing the nature of PRMs. 
% In addition, using PRMs generated in the static environment for dynamic scenarios could be problematic.
% A connected graph with vertices sampling only in static environments cannot guarantee the feasibility of the roadmap in the temporal domain, as shown in Fig. \ref{fig:valid_graph}.

% Previous works have focused on utilizing sampled valid vertices with time stamps in dynamic environments, However, employing such a strategy with a finite number of sampling vertices can be highly inefficient in this setting, leading to redundant edge traversals.
% Other works also consider sample valid vertices with safe time intervals (temporal intervals that the vertices are collision-free within), while limiting the maximum length of edges for efficiency. 

The algorithm for constructing the graph is detailed in Algorithm~\ref{alg: graph}. 
Within a dynamic connected visibility graph, vertices are classified as either Guards or Connectors. 
% Vertices are classified as Guard or Connectors depending on whether they can form edges with any other two vertices.
Vertices are classified as Connectors if they can connect with any other two Guards; otherwise, they are classified as Guards.
% Guards denote vertices incapable of connecting to any other two Guards, while Connectors are those capable of connecting to at least two Guards (see Fig. \ref{fig:valid_graph}). 

The start and goal vertices are initiated as Guards($Guard()$) into the graph. 
In the main loop, \textbf{getSample()} function employs a heuristic strategy to sample vertices uniformly from regions with lower Connector-Guard ratios.
The \textbf{findVisibleGuard()} identifies Guards $g_1, g_2$ that can connect to a sampled vertex $v$.
Safe intervals are determined for the edge connecting $g_1, v$ and the edge connecting $g_2, v$. They can be connected if any of these safe intervals overlap.
Neighbors for $g_1, g_2$ are then identified by \textbf{neighbors()}.
As the connection direction is ambiguous, both the forward path $\overrightarrow{\varsigma_1}$ and the reverse path $\overleftarrow{\varsigma_1}$ are considered. 
UTVD class is checked between newly sampled path and neighbor paths by \textbf{checkEquiv()}, each path within their safe intervals is discretized into a set of points with time stamps, and then UTVD is employed to assess equivalence.
If they belong to the same class and the newly sampled path is shorter, the neighbor vertex is replaced with the sampled vertex. 
Otherwise, the sampled vertex is designated as a Connector($Connector()$), generating new edges for both $g_1$ and $g_2$ in \textbf{addNewEdge()}.
This dynamically connected visibility graph meets the dynamic feasibility
requirements and guarantees collision-free edges, allowing for
various parameterizations along each edge. 
For instance, the quadrotor can smoothly accelerate, decelerate, and stop along edges while avoiding collisions within safe intervals.
\vspace{-0.2cm}
\begin{algorithm}
	\caption{Dynamic Connected Visibility Graph}
	\label{alg: graph}
    	\KwIn{start position $x_s$, goal position $x_g$, time range $T$, Obstacle information $\mathcal{X}_{obs}$}
        \KwOut{graph $G$}
        $G \leftarrow \varnothing $,\\
        $G \cup Guard(x_s) \cup Guard(x_g)$, \\
        \Repeat{end condition}
        {
            $v$ = \textbf{getSample}() \\
            $guards$ = \textbf{findVisibleGuard}($v, \mathcal{X}_{obs}, G, T$)\\
            \If{$guards = \varnothing $}
            {
                $G \cup Guard(v)$ \\
                continue\\
            }
            \If{$guards$ has two vertices $g_1, g_2$}
            {

                $\varsigma_1 \leftarrow (g_1, v ,g_2$) \\
                ${\rm SI}(g_1, v) \leftarrow $ {\rm \textbf{findSafeIntervals($g_1, v$) }} \\
                ${\rm SI}(v ,g_2) \leftarrow $ {\rm \textbf{findSafeIntervals($v ,g_2$) }} \\
                \If { ${\rm SI}(g_1, v) \cap  {\rm SI}(v ,g_2) =  \varnothing $         }
                {
                    continue
                }

                \For{each $n_g \in $ \textnormal{\textbf{neighbors}($g_1, g_2$)}}
                {   
                    $\varsigma _2 \leftarrow (g_1 , n_g , g_2)$ \\
                    isSameTopo = \textbf{checkEquiv}($\overrightarrow{\varsigma _1}$, $\overrightarrow{\varsigma _2}$) $\vee$\textbf{checkEquiv}($\overleftarrow{\varsigma _1}$, $\overleftarrow{\varsigma _2}$) \\
                    
                    \If{ {\rm isSameTopo}$\wedge$\rm{$\| \varsigma _1\|^2<\|\varsigma _2 \|^2 $} }
                    {
                        $n_g \leftarrow v$\\
                        break\\
                    }
                }
                \If{$\neg$ {\rm isSameTopo} }
                {
                    $G \cup Connector(v, {\rm SI}(g_1, v), {\rm SI}(v ,g_2))$ \\
                    \textbf{addNewEdge($v, g_1, g_2$)}
                }
            }
        }

\end{algorithm}
\vspace{-0.5cm}
Given a graph, the depth-first search algorithm is employed to explore multiple distinct topological paths.
For each path, vertex parameterization by time is implemented by the rule similar to \cite{SIPP1} (i.e., reach vertices as early as possible). 
The initial temporal corridors are constructed by Definition \ref{def: tempcorr}.

% Specifically, for a given vertex, the overlap intervals of the safe intervals of the edges connecting to it are calculated.
% The earliest reachable time for this vertex is feasible if the robot traveling from the previous vertex can reach this vertex at the time within the overlap intervals.
% Once the parameterized paths are identified, they can be utilized for subsequent trajectory optimization. 

\subsection{Theoretical Analysis}

With the assumption that the environment can be fully represented by graphs, we further discuss the optimality and completeness of the proposed approach.
SIPP-based methods discretize an environment into grids with the subsequent composition of vertices within a graph, and the time domain is continuously evaluated. 
The SIPP is proven to be complete and resolution-optimal\cite{SIPP1} without dynamic constraints and action availability. 
Safe Interval-based methods are intrinsically equivalent to finding a feasible temporal corridor. 
By introducing PRM-based strategies and dynamic considerations, our proposed method can achieve probabilistic complete and optimal path planning within a finite duration.

\section{Trajectory Planning with Spatial-Temporal Corridors}

The front-end method provides the initial paths in distinct UTVD classes and corresponding temporal corridors. 
To further ensure the smoothness of the third-order systems, We apply the uniform B-spline curve to represent the trajectory, with the advantages of its convex hull property to enforce dynamic feasibility and geometric constraints.

\vspace{-0.2cm}
\subsection{Trajectory Optimization Formulation}
\vspace{-0.2cm}
We can efficiently parameterize the continuous trajectory in its flat space because of differential flatness\cite{5980409}.
Given a collision-free path generated considering lower-order dynamics $\Gamma :[t_s, t_e] \ \mapsto \in \mathbb{R}^3 $, we can generate a n-dimension $p_b$ degree uniform B-spline constructed by control points $\mathbf{Q} = [\mathbf{Q}_1, \cdots, \mathbf{Q}_{N_c} ]^T, \mathbf{Q}_i \in \mathbb{R}^3 $, and a knot vector $\mathbf{t} = [t_1, \cdots, t_{N_c+p_b} ]^T \in  \mathbb{R}^{N_c+p_b} $ with identical knot span $t_s$. 
Hence, we formulate the optimization problem as
\vspace{-0.2cm}
\begin{align}
        \min_{ \bm{Q, t} } \  \sum_d^D \lambda_d J_d(\bm{Q, t}),
\end{align}
where $D = \{ c, od, ct, f \}$, $J$ represents the control cost (c), the collision cost with moving obstacles (od), the spatial-temporal corridor cost (ct), and the dynamic feasibility cost (f), and $\lambda$ denotes the corresponding weights.
We adopt a framework similar to that in~\cite{fasttopo, 9309347} for optimizing control points with fixed time knots and iteratively refining time allocation.
The one with minimum control cost is selected among multiple trajectories.

\vspace{-0.2cm}
\subsection{Spatial-Temporal Corridor Inflation}
\begin{figure}[!ht]
      \centering
      \vspace{-0.1cm}
      \subfigure[1-D example]{
      \includegraphics[width=0.46\columnwidth]{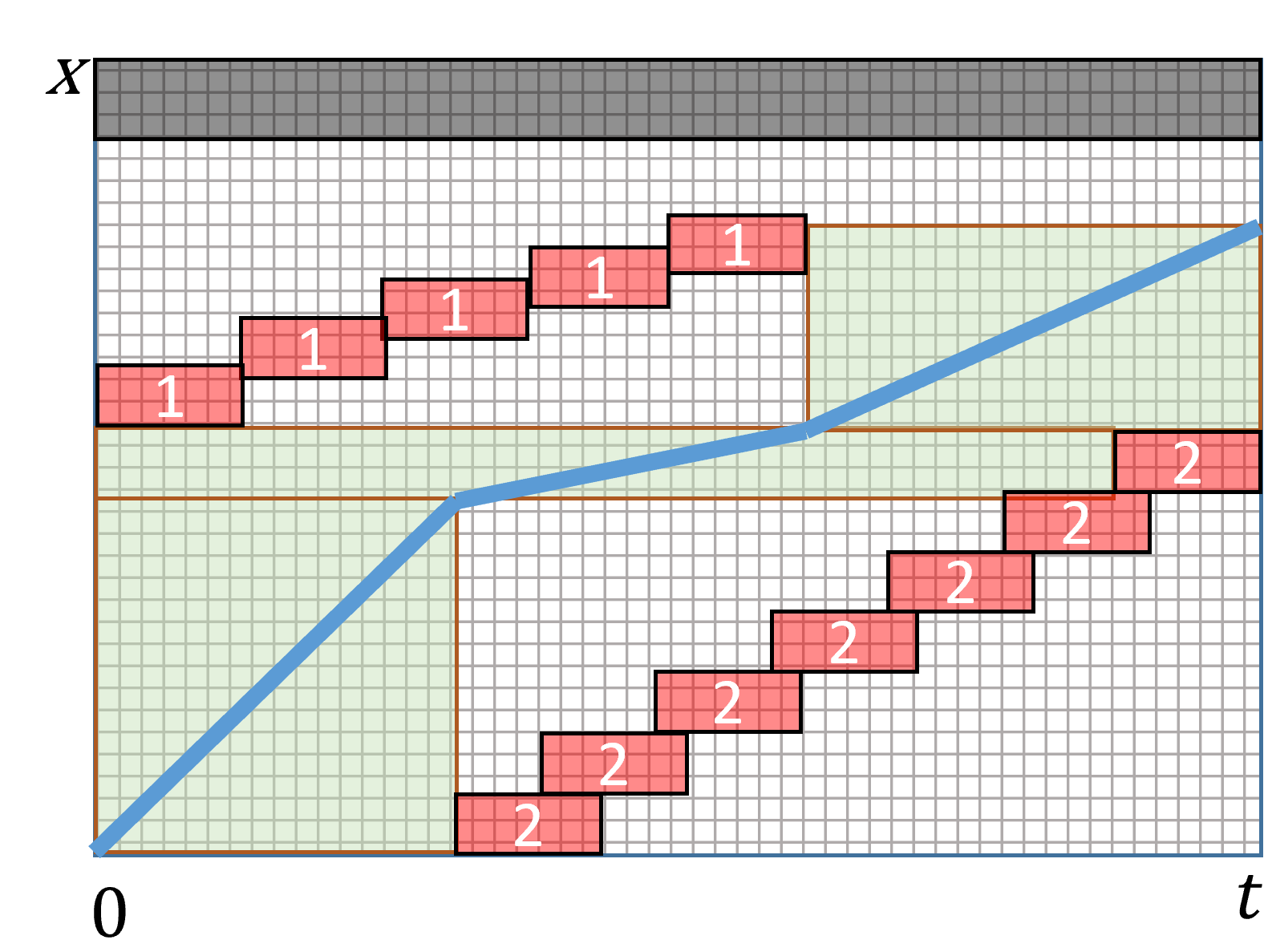}
      \label{fig:infla1}
      }
      \subfigure[Spatial-temporal corridor inflation]{
      \includegraphics[width=0.46\columnwidth]{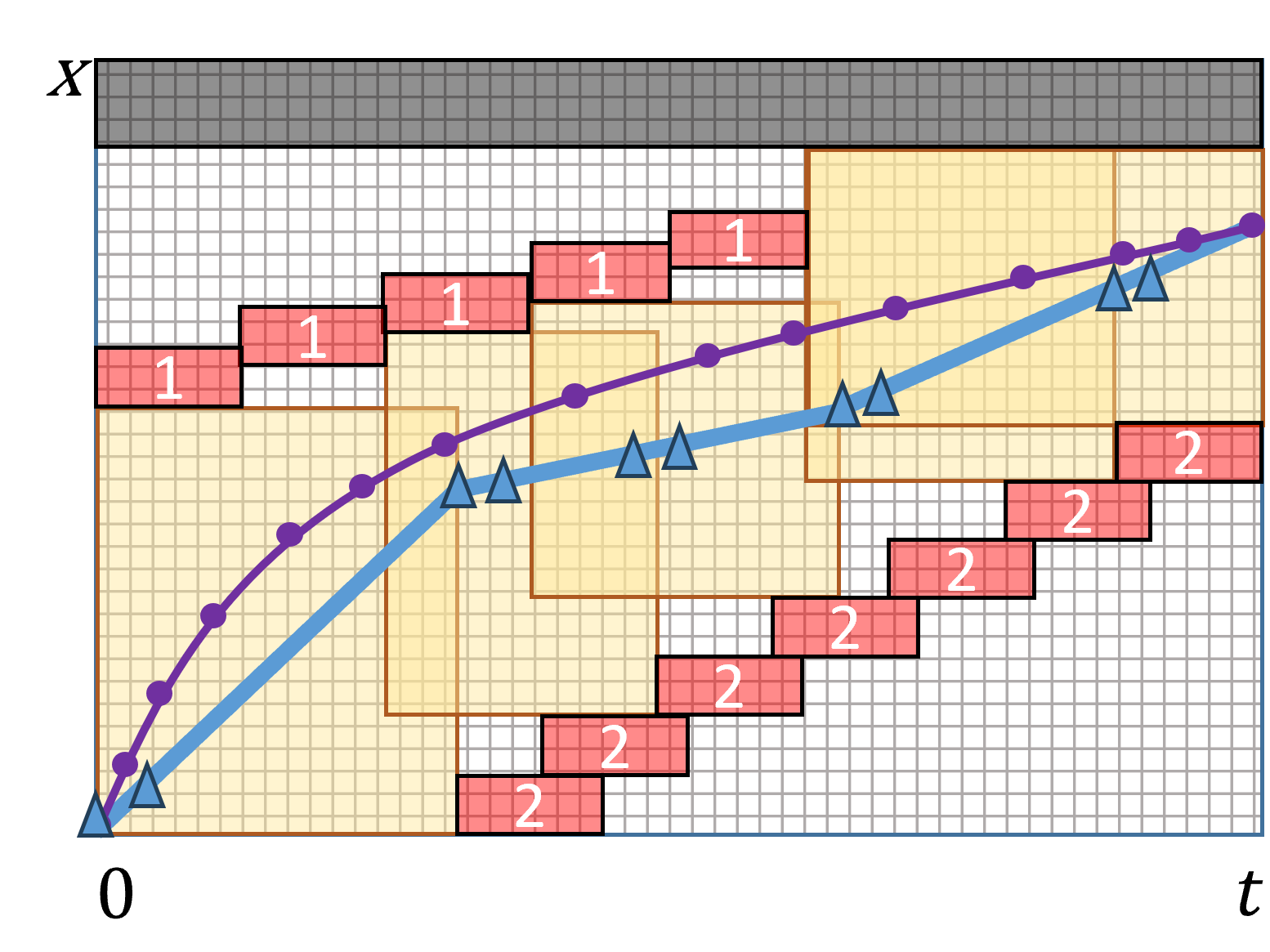}
      \label{fig:infla2}
      }
      \vspace{-0.5cm}
      \caption{Demonstration of spatial-temporal corridors inflation in one dimension, (a) denoted by position (x) versus time (t). The initial path and initial spatial-temporal corridors are represented in blue and green, static obstacles are shown in black while moving obstacles 1 and 2 are colored in red. (b) The inflated spatial-temporal corridors are shown in yellow, which are generated by seed points in triangular shapes. The optimized B-spline trajectory with control points is colored in purple.}
       \label{fig:infla}
       \vspace{-0.5cm}
\end{figure}

\vspace{-0.1cm}
% The spatial-temporal corridors generated around the initial paths only cover the edges of the geometric space, limiting the trajectory's flexibility to further adjust around these paths.
The front-end path is an initial spatial-temporal corridor, where edges define spatial corridors and time intervals set temporal corridors. However, the spatial corridors are confined to geometric edges, limiting trajectory flexibility.
Hence, we need to inflate both the spatial and temporal corridors in 4-D space~\cite{ding2019safe}.
Specifically, for each edge, we incrementally select two successive seed points along it, inflating the spatial corridors as axis-aligned cuboids in 3-D space while inflating the temporal corridors by checking potential collisions in these cuboids.
Each spatial-temporal corridor is defined by a feasible temporal interval $(t_l, t_u) $ within which the cuboid is collision-free, and boundary points of the cuboid $\{b_l, b_u\}$, where $b_l$ and $b_u$ are the lower bound and upper bound in each 3-D dimension, respectively. Fig. \ref{fig:infla} provides a 1-D example of spatial-temporal corridor inflation.
% \begin{figure}[!ht]
%     \centering
%     % \subfigure[Static connected graph]{
%     %   \includegraphics[width=0.46\columnwidth]{fig/grapha.png}
%     %   \label{fig:valid_graph}
%     % }
%     %\subfigure[Dynamic connected graph]{
%       \includegraphics[width=0.5\columnwidth]{fig/graphb.png}
%       \label{fig:valid_graph2}
%     %}
%     \caption{Illustration of dynamically connected visibility PRM in a dynamic environment. The dynamic connected graph has edges that are valid in their safe intervals.}
%  \label{fig:approximation}
% \end{figure}
\vspace{-0.3cm}

\subsection{Objectives Evaluation}

%A uniform $p_b$ degree B-spline curve can be represented simply by matrix calculation of knot vector, control points and a constant matrix related to degree of B-spline~\cite{731996}. 
We derive control points for higher-order velocity, acceleration, and jerk to evaluate the minimum control and dynamics cost of the B-spline trajectory.
We represent the velocity control points $\mathbf{V} = [\mathbf{V}_1, \cdots, \mathbf{V}_{N_c-1} ]^T$,  acceleration control points $\mathbf{A} = [\mathbf{A}_1, \cdots, \mathbf{A}_{N_c-2} ]^T$ and jerk control points $\mathbf{J} = [\mathbf{J}_1, \cdots, \mathbf{J}_{N_c-3} ]^T$ using control points $\mathbf{Q}$ and knot span $t_s$, as
\vspace{-0.1cm}
\begin{gather}
    \mathbf{V}_i = \frac{\mathbf{Q}_{i+1} - \mathbf{Q}_{i}}{t_s}, \quad  
    \mathbf{A}_i = \frac{\mathbf{V}_{i+1} - \mathbf{V}_{i}}{t_s}, \quad 
    \mathbf{J}_i = \frac{\mathbf{A}_{i+1} - \mathbf{A}_{i}}{t_s}.
\end{gather}
The control cost function $J_c$ is also formulated as penalizing the jerk of the trajectory,
\vspace{-0.2cm}
\begin{equation}
       J_c = \sum_{i=p_b-3}^{N_c-p_b}\|\mathbf{J}_i\|_2^2 .
\end{equation}
The dynamically feasible cost is formulated to penalize the trajectory with exceeding maximum velocity $v_m$ and maximum acceleration $a_m$ with respect to each dimension:
\vspace{-0.25cm}
\begin{equation}
       J_f = \sum_{i=p_b-1}^{N_c-p_b}\|\mathbf{V}_{i} - v_{m}\|_2^2 + 
             \sum_{i=p_b-2}^{N_c-p_b}\|\mathbf{A}_{i} - a_{m}\|_2^2 . 
\end{equation}
% by the convex hull property of B-spline, dynamic feasibility can be enforced if all first and second-order derivatives of control points are in the feasible region.
%We define the collision cost with dynamic obstacles as 
We also incorporate a collision cost for dynamic obstacles, as the generated trajectory may come too close to the boundaries of cuboids.
\vspace{-0.2cm}
\begin{gather}
    J_{od} = \sum_{i=0}^m\sum_{j=0}^k J_{ij}, \\ 
       J_{ij} = \left\{
        \begin{array}{ll}
            0 & \text{if } d(p_i, o_{j}) > d_{th} \\
            (d(p_i, o_{j}) - dth)^2 & \text{if } d(p_i, o_{j}) \leq d_{th}
        \end{array} \right.,
\end{gather}
where $m$ is the number of samples, $k$ is the number of dynamic obstacles.
We applied the Euclidean distance function, $ d(p_i, o_j) = \|E_j^{-1}(p_i - o_{j})\|_2 $ between $i^{th}$ sampled point and $j^{th}$ dynamic obstacle center, $E_j$ is the coefficient matrix of an ellipsoid-shaped moving obstacle, $d_{th}$ is the minimum distance threshold.

The spatial-temporal corridor cost is defined as the L1 norm of the distance between control points and cuboids,
\vspace{-0.2cm}
\begin{equation}
       J_{ct} = \sum_{i=p_b}^{N_c-p_b}(\|b_{l,j} - \mathbf{Q}_{i}\|_1 + \|\mathbf{Q}_{i} - b_{u, j}\|_1).
\end{equation}
 We identify the cuboid corresponding to $\mathbf{Q}_{i}$ by locating the spatial corridor whose temporal interval includes the knot of $\mathbf{Q}_{i}$.

% Fig. \ref{fig: traj} shows the 3-D simulation of back-end optimization with spatial-temporal corridors. The top-down view depicts a quadrotor's trajectory from 0s to 9s. 
% The initial B-spline trajectory is blue, corridors cubes are yellow, moving obstacles are in red while static obstacles are in black, each moving obstacle within the scope is attached with an id, and green cycles highlight the quadrotor's position.
% The final trajectory with minimum control cost is represented in pink. 

\begin{figure}[!ht]
    \centering
    \includegraphics[width=1.0\columnwidth]{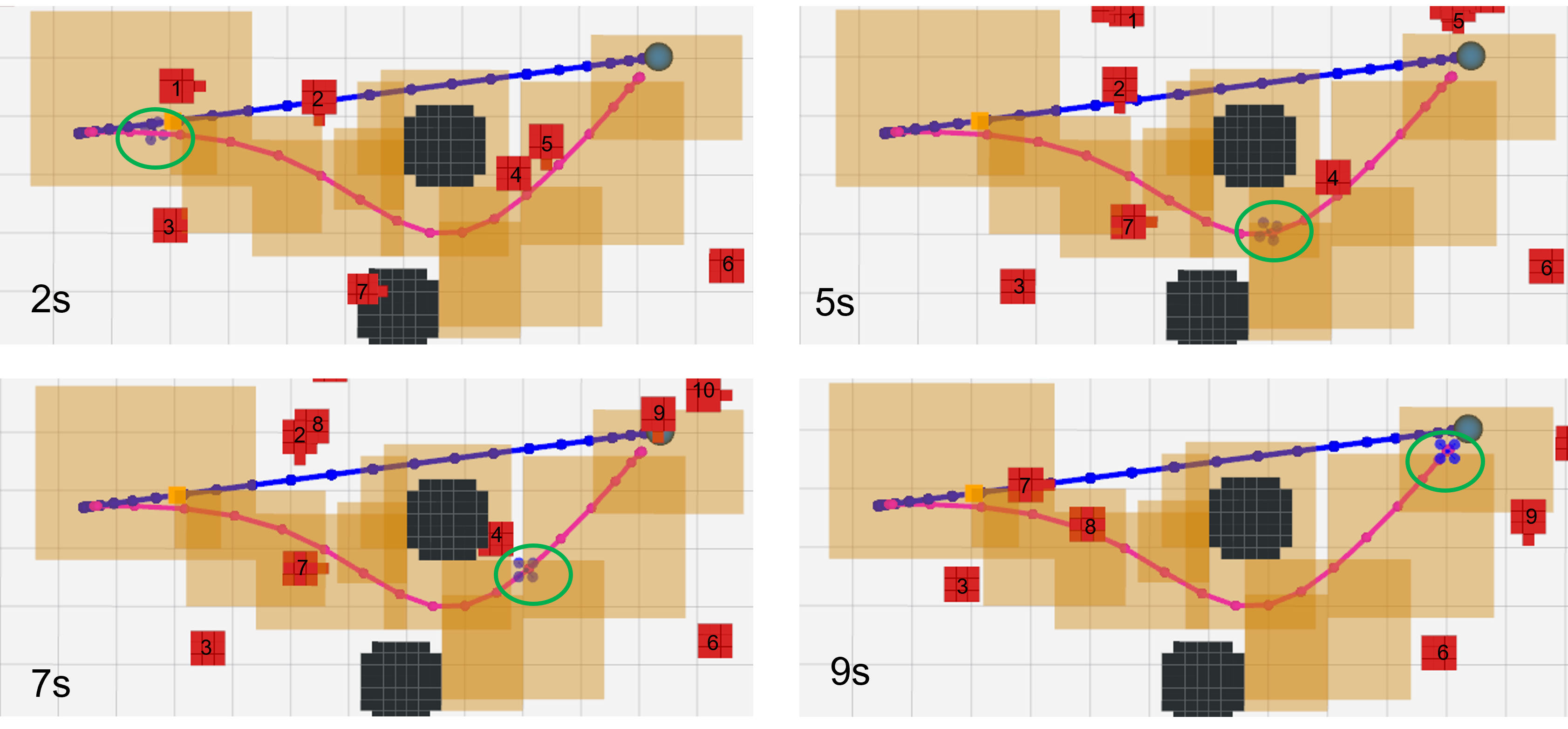}
    \vspace{-0.7cm}
    \caption{Simulation result (top-down view in 3-D environment) of the back-end optimization. Each cuboid is valid within its temporal intervals. The yellow cuboid demonstrates the spatial-temporal corridor.}
    \vspace{-0.7cm}
    \label{fig: traj}
\end{figure}

Fig. \ref{fig: traj} visualizes back-end optimization with spatial-temporal corridors. The top-down view shows a quadrotor’s trajectory (0s–9s), the initial B-spline (blue), corridor cuboids (yellow), moving obstacles (red) with IDs, static obstacles (black), and the quadrotor’s position (green circles). The optimized trajectory with minimal control cost is in pink.

\section{Results}

\begin{figure*}[!ht]
      \vspace{-0.2cm}
      \centering
      
      \includegraphics[width=2.0\columnwidth]{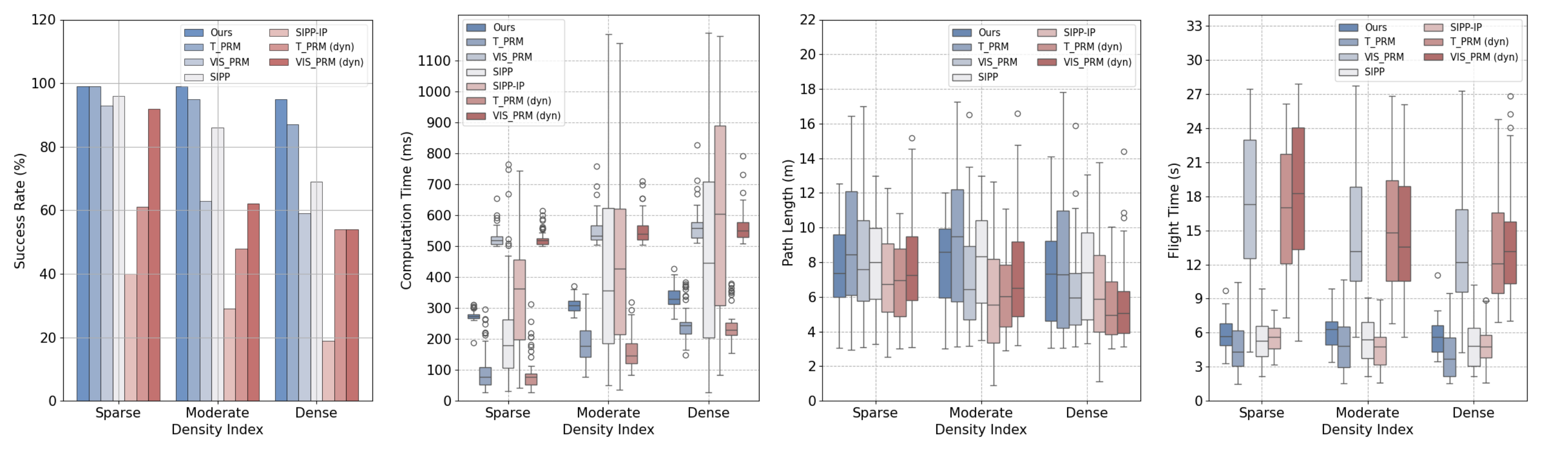}
      \vspace{-0.5cm}
      \caption{Simulation results for front-end methods for 100 trials in three different complexity levels maps. Density is calculated as the ratio of occupied grid cells to total grid cells in the bounded space. The density range of sparse, moderate, and dense maps are $[0, 0.01]$, $[0.05, 0.1]$, and $[0.15, 0.2]$, and the range of number of dynamic obstacles are $[0, 20]$, $[20,  40]$, and $[40, 60]$, respectively. }
      \label{fig: bench_fig}
      \vspace{-0.5cm}
\end{figure*}

\subsection{Implementation Details}
\vspace{-0.1cm}
We apply the parameterized environment generation in \cite{10610207} and extend to moving obstacles with varying ellipsoidal sizes and minimum acceleration trajectories in three types of maps with different levels of complexity. 
% The evaluation is conducted on the 3-D map with the size of $20 \times 20 \times 5$ m$^3$ and resolution as $0.2$ m, in different complexity levels. 
% creating multi-shaped obstacles stored within the occupied map with the size of $20 \times 20 \times 5$ m$^3$ and resolution as $0.2$ m.
% For dynamic moving obstacles, an extension of this approach involves randomly generating obstacles with varying ellipsoidal sizes. 
%Initially, each obstacle is randomly assigned two positions, and an obstacle prediction horizon is set to $25$ seconds. 
%Within each horizon, a minimum-acceleration planner generates trajectories for each obstacle so that it can move between the two positions. 
% The trajectories are generated and broadcast to the robot. 
% The obstacle-bounded velocity and acceleration in each dimension are set to $0.25m/s$ and $0.25m/s^2$, respectively.
% The maximum magnitude velocity and acceleration of the robot are limited to $1.5 m/s$ and $1.5 m/s^2$.
The collision-free start and goal positions are randomly generated for each trial, and each map is re-generated randomly every three trials. 
The plan is considered successful if the quadrotor can find a path or trajectory from the start to the goal position without collision.
\vspace{-0.2cm}
\subsection{Simulation Experiments}
\subsubsection{Front-end Paths Benchmarks}

We compared the performance of our front-end approach with four different path planners: T\_PRM~\cite{TPRM}, VIS\_PRM~\cite{topo1}, SIPP~\cite{SIPP1}, and SIPP\_IP~\cite{kinoSIPP}.
%across 100 trials in different complexity level of 3-D maps filled with both static and dynamic obstacles. 
To demonstrate dynamical feasibility, we incorporate kinodynamic checking into VIS\_PRM and T\_PRM by determining the feasible traversal time using a trapezoidal velocity profile instead of assuming a constant maximum speed, denoted as ``(dyn)".
%We classify the density index from Environment Complexity Signature \cite{10610207} into three levels together with the number of moving obstacles to parameterize the dynamic environment. 
We evaluate these methods with respect to success rate,  computational time, path length, and flight time.
% , and arrival time.
The safety for success rate computation is evaluated by checking if edges are collision-free within the duration of start and end vertices. 
As shown in Fig. \ref{fig: bench_fig}, our method achieves a higher success rate in all environments of different density levels.
% good path quality performance when considering dynamical feasibility.
In addition, the proposed method achieves relatively low flight times in all cases. 
%%%% give a summary.

% For front-end methods without dynamic considerations, SIPP has a large drop when the environment becomes more dense.
% SIPP-based methods reply on the time intervals precomputation of each grid, the computation time is much longer than others. 
SIPP-based methods, which rely on pre-computing time intervals for each grid, experience significantly longer computation times and performance degradation in denser environments.
T\_PRM demonstrates a high success rate and low computation time, with slightly longer path lengths than our methods.
As collision avoidance of T\_PRM is guaranteed by checking the vertices, the edge lengths should be constrained into a reasonable range to prevent unsafe connection, which results in more convoluted paths. 
VIS\_PRM samples vertices with timestamps and connects them if the edge remains collision-free during a defined period. 
% While both methods produce similar average path lengths, VIS\_PRM 
It has a lower success rate as moving obstacles increase, compared with methods that leverage the completeness %efficiency 
of safe interval planning.

%We integrate dynamic feasibility checking to VIS\_PRM and T\_PRM by determining the feasible traversal time using a trapezoidal velocity profile instead of assuming a constant maximum speed. 
The success rate of T\_PRM (dyn) decreases, and flight times significantly increase compared to the original T\_PRM, showing that the maximum edge length alone cannot ensure complete safety when dynamic feasibility is taken into account. 
%Additionally, flight times and arrival times experience significant increases compared to the original method, highlighting that our approach yields much shorter path duration even when factoring in dynamic feasibility constraints.
VIS\_PRM (dyn) doesn't demonstrate a significant change compared to the original VIS\_PRM, because random sampling of vertex time stamps usually yields sufficient time durations to ensure dynamic feasibility.
% The SIPP\_IP extended the original SIPP by generating a motion primitive set and project waiting intervals along these motion primitives.
% SIPP\_IP has the longest computational time and often terminates upon reaching the maximum number of expanded vertices.
% Two potential factors contribute to this outcome: Firstly, generating motion primitives in a 3-D grid world is inherently slow. 
% Secondly, because the heuristic function does not consider dynamic feasibility, the search process is sluggish, leading to a success rate of less than 50\%.
The SIPP\_IP algorithm has the longest computational time and often stops after reaching the maximum number of expanded vertices.
%The SIPP\_IP algorithm, which generates motion primitives and projects waiting intervals, has the longest computational time and often stops after reaching the maximum number of expanded vertices.
%It's due to the slow generation of 3D motion primitives and a heuristic that doesn't consider dynamic feasibility, leading to a success rate of less than 50\%.

% Note that our method shows relatively short flight and arrival times but has more outliers than other methods with the same flight and arrival times.
% We consider safe intervals of edges, which might not start immediately, even if the starting point is safe, which allows the robot to wait until the first safe interval begins, assuming safety during the waiting period ensured by back-end optimization. 
% In addition, we apply a conservative dynamic feasibility, which results in longer traversal times on edges, causing longer flight times compared to methods that assume constant maximum speed.

\subsubsection{Trajectory Evaluations}

To further validate the effectiveness of our framework, we evaluate our planner by comparing its performance with a baseline planner: TPRMO, which uses TPRM as the front-end method and optimizes it without the use of spatial-temporal corridors.
The front-end and back-end planning framework is triggered when a collision is detected on an initial B-spline trajectory. 
%We only evaluate trajectories that include valid front-end paths.
Trajectories are generated and evaluated if the front-end planner can find a valid path.
For the baseline method, we include the Euclidean distances towards moving obstacles as cost functions and use the static obstacle avoidance strategy mentioned in \cite{9309347}.
We conducted 100 trials for each type of map.
% The results in Table \ref{table1} show that the method with TC significantly improves the success rate in one-time optimization (without re-planning upon failure). 
% In addition, the average trajectory length and duration are reduced in all three cases. 
% It demonstrates that using the temporal corridor method can produce less conservative trajectories.
% Re-parameterizing trajectory time within the safe temporal intervals ensures both the safety and efficiency of the proposed method.
The results in Table \ref{table1} show that the success rate in one-time optimization (without re-optimizing upon failure), the average trajectory length, and duration are comparable between the two methods. However, the average control cost (integral of the squared jerk) is significantly reduced using our method. 
It demonstrates that our two-stage method, utilizing the spatial-temporal corridor, produces smoother trajectories.
Re-parameterizing trajectory time within the corridor ensures both safety and efficiency.

\begin{table}
\centering
\caption{Planner comparison}
\vspace{-0.2cm}
\label{table1}
\begin{tabular}{ |p{1cm}|c|c|c|c|c| }
\hline
Env. & Methods & \tabincell{c}{Opt.\\Succ.\\ Rate}  & \tabincell{c}{Traj. \\ Len. (m)} & \tabincell{c}{Flight \\ Time (s)}  & \tabincell{c}{Ctrl. Cost \\ Avg. (m$^2$/s$^5$)}\\
\hline
\multirow{2}{4em}{Sparse}
& Ours & 100\% & 7.66  & 6.04 & \textbf{17.71}\\ 
& TPRMO & 99\% & 7.89 & 5.55 & 72.67\\ 

\hline
\multirow{2}{4em}{Moderate}
&  Ours & 98\% & 7.93  & 6.05 & \textbf{26.20}\\ 
& TPRMO& 97\% & 8.11 & 5.72 & 73.27\\ 

\hline
\multirow{2}{4em}{Dense}
& Ours & 97\% & 7.44  & 5.69 & \textbf{30.53}\\ 
& TPRMO & 97\% & 7.52 & 5.36 & 70.31\\
\hline
\end{tabular}
\vspace{-0.8cm}
\end{table}

\subsection{Hardware Experiments}
\begin{figure}[!t]
    \centering
    \includegraphics[width=1.0\columnwidth]{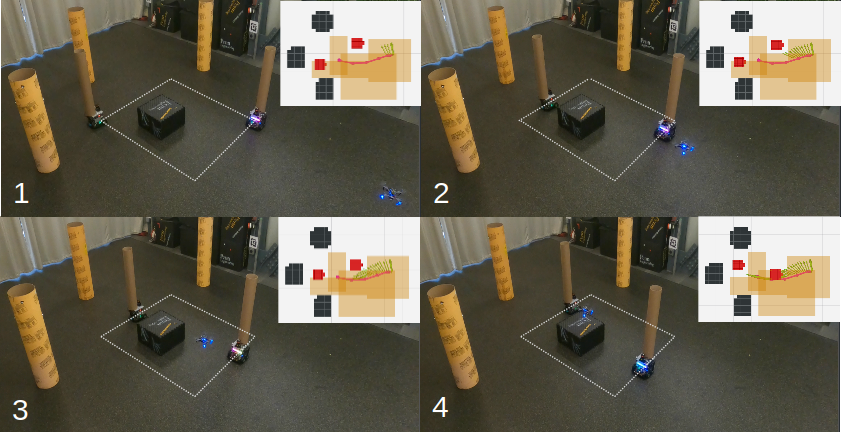}
    \vspace{-0.5cm}
    \caption{Real-world experiment with two moving obstacles and three static obstacles. In the upper-right visualization in the top-down view, the red and black polygons correspond to the moving obstacles and static obstacles. Besides, the red curve and green arrows represent the quadrotor trajectory and odometry, respectively. The experiment video is available at \url{https://youtu.be/Bx_q_11eOrg}.}
    \vspace{-0.8cm}
    \label{fig: real-world experiment}
\end{figure}

\vspace{-0.1cm}
We validated our proposed framework with extensive real-world experiments. 
% We conducted extensive real-world experiments to validate the effectiveness of the proposed framework. 
To create moving obstacles, we employed two Scarab ground robots~\cite{Scarab2008}. Each Scarab is equipped with a Hokuyo UTM30LX laser and an onboard computer with an Intel i7-8700K CPU. In addition, we mount a $0.91m \times 0.16m$ cylinder on each of them. We set up static obstacles with three $1.2m \times 0.3m$ cylinders. 
A customized Dragonfly 230 quadrotor is used to carry out the experiments. It carries a VOXL flight board, a forward-facing Time-of-Flight camera, and a downward-facing tracking camera, as detailed in work~\cite{10.1007/978-3-030-71151-1_37}.
The Vicon Motion Capture system is used to set up the common reference frame and provide odometry information. 
One set of our experiments is demonstrated in Fig.~\ref{fig: real-world experiment}.
The quadrotor first took off from the bottom-right corner while the scarab robots were tracking a rectangle trajectory, 
%a constant velocity at $0.xm/s$,
as illustrated in white dotted lines. A navigation goal centered on top of the black box was set and the planner was triggered. As shown in the upper-right corner, the spatial-temporal corridor was built and a minimum control cost trajectory was generated. The quadrotor tracked the planned trajectory closely and reached the navigation goal, as illustrated in the sub-figures. Subsequent navigation goals were set after the quadrotor reached the first goal.
%. In these experiments, we limit the max velocity and acceleration of the quadrotor at $1.0m/s$ and $1.0m/s^2$.
\vspace{-0.1cm}
\section{Conclusion} 
\label{sec:conclusion}
\vspace{-0.1cm}
This paper addresses the dynamic obstacle avoidance problem and introduces a complete two-stage planning approach to efficiently identify the feasibility of the environment setting and generate smooth trajectories. 
We apply a front-end graph construction and search method to identify multiple distinct paths in different spatial-temporal topological classes based on the concept of UTVD.
Spatial-temporal corridors are subsequently constructed to optimize B-spline trajectories, ensuring safety, dynamical feasibility, and smoothness in environments filled with both static and moving obstacles. 
For future work, we plan to integrate the proposed method with onboard perception systems to achieve more robust and reliable performance in dynamic environments featuring obstacles with complex movement patterns.

\bibliography{references}
\end{document}